\documentclass{article}
\usepackage{log_2025}						

\usepackage{booktabs}						
\usepackage{multirow}						
\usepackage{amsfonts}						
\usepackage{graphicx}						
\usepackage{duckuments}						

\usepackage[numbers,compress,sort]{natbib}	

\usepackage{wrapfig}
\usepackage{float}

\title[Learning Particle Dynamics Subject to Rigid Body Manipulations Using Graph Neural Networks]{Learning Particle Dynamics Subject to Rigid Body Manipulations Using Graph Neural Networks}

\author[Midlagajni et al.]{%
 Niteesh Midlagajni \\
  Centre for Cognitive Science\\
  Technical University of Darmstadt\\
  \texttt{niteesh.midlagajni@tu-darmstadt.de} \\
  \And
   Constantin A. Rothkopf \\
  Centre for Cognitive Science\\
  Technical University of Darmstadt\\
  \texttt{constantin.rothkopf@tu-darmstadt.de} \\
}

\begin{document}

\maketitle

\begin{abstract}
Simulating particle dynamics with high fidelity is crucial for solving real-world interaction and control tasks involving liquids in design, graphics, and robotics. Recently, data-driven approaches, particularly those based on graph neural networks (GNNs), have shown progress in tackling such problems. However, these approaches are often limited to learning fluid behavior in static free-fall environments or simple manipulation settings involving primitive objects, often overlooking complex interactions with dynamically moving kinematic rigid bodies. Here, we propose a GNN-based framework designed from the ground up to learn the dynamics of liquids under rigid body interactions and active manipulations, where particles are represented as graph nodes and particle-object collisions are handled using surface representations with the bounding volume hierarchy (BVH) algorithm. Our approach accurately captures fluid behavior in dynamic settings and can also function as a simulator in static free-fall environments. Despite being trained on single-object manipulation tasks, our model generalizes effectively to environments with novel objects and novel manipulation tasks. Finally, we show that the learned dynamics can be leveraged to solve control and manipulation tasks using gradient-based optimization methods.
\end{abstract}    
\section{Introduction}
\label{sec:intro}

Our physical world is shaped by complex interactions between various forms of matter, from solid objects to flowing liquids. Accurately simulating these phenomena is essential across numerous disciplines, including science and engineering, and becomes increasingly critical as we advance toward a future with autonomous agents capable of reasoning, planning, and interacting with objects in novel environments. 
Moreover, understanding human intuitive physical reasoning and interactions in everyday tasks in the natural environment also requires the capability of simulating physical scenarios \cite{intuitive_liq, intuitive_phy}. Traditional physics simulators, relying on hand-crafted features and computationally intensive numerical methods, often struggle to scale to the complexity of real-world scenarios.

\begin{figure}[t]
    \centering
    \includegraphics[width=\linewidth]{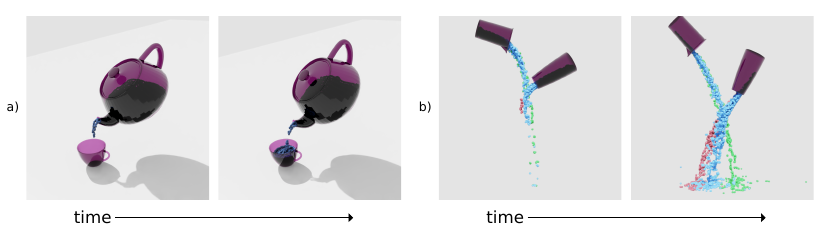}
    \caption{Rollouts from our proposed model on novel scenarios. (a) Successful simulation of pouring from the complex Utah teapot. (b) Simultaneous manipulation of two jugs, demonstrating multi-object control. The blue liquid streams, originating from each jug, collide and merge realistically. The red and green particles represent the predicted trajectories for each jug if simulated independently.}
    \label{fig:2jugs_tea}
\end{figure}

Recently, learning-based simulators have offered a viable alternative, demonstrating the ability  to learn complex dynamics directly from data \cite{he2019learning, wiewel2019latent, ladicky2015data}, improve simulation accuracy, and enable natural integration with gradient-based optimization for solving inverse problems \cite{challapalli2021inverse, lin2022diffskill}. Graph neural networks (GNNs) \cite{gnn_theory_2,gnn_theory}, in particular, have been successful in recent years in capturing the dynamics of diverse physical systems, including liquids, soft bodies, and rigid bodies \cite{sanchez2020learning,meshgraphnet,fignet,sdf_learning,SURFsUP,dpinet}. While existing GNN-based simulators have shown impressive results in modeling particle dynamics, significant limitations remain.  Many current approaches \cite{dpinet,sanchez2020learning} are restricted to simple collision scenarios with basic geometric shapes, hindering their applicability to real-world rigid body interactions. While recently, \citet{SURFsUP} improved boundary handling with complex rigid bodies via Signed Distance Fields (SDFs), the model is still constrained to free-fall simulations, where the rigid bodies remain fixed in place during simulation. Accurately modeling dynamic liquid interactions with moving objects (kinematic rigid bodies) is essential for applications in sequential decision-making tasks such as robotic control. A simulator designed for free-fall settings can address only design-oriented tasks (e.g., directing fluid flow \cite{inverse_design}), and is insufficient for tasks requiring manipulation and control. Crucially, even with highly accurate collision handling, a static framework does not guarantee correct behavior when objects move, as dynamic interactions introduce additional complexities in fluid motion and response. Therefore, achieving accurate dynamic simulations requires careful design choices, including the appropriate GNN input representation and architectural modifications.

To address these challenges, we propose a unified GNN-based framework that models complex fluid dynamics with dynamically moving rigid bodies. Our architecture leverages a multi-graph representation where liquid particles and rigid bodies are represented as distinct node sets. Liquid–liquid interactions are determined by spatial proximity (particles are connected if they lie within a fixed radius), while liquid–object interactions are determined by particle–surface proximity computed efficiently using the BVH algorithm. We train our model on ground truth simulations of generic \textit{pouring} and \textit{sliding} tasks, collectively covering a broad range of liquid-object interactions under dynamic control. Our results show: 1) improved performance both qualitatively and quantitatively over existing GNN-based particle simulators; 2) strong generalisation to unseen object geometries and scene configurations; and 3) successful zero-shot transfer to novel manipulation tasks such as \textit{stirring} and \textit{scooping}. Finally, we demonstrate the utility of our model by integrating it into a gradient-based optimization pipeline to solve a pouring task using a Model Predictive Control (MPC) controller. Taken together, these results demonstrate that our proposed model extends the capabilities of GNN-based simulators to include complex interactions of liquids with kinematic rigid bodies.

\section{Related Work}
\label{sec:related}

With the rise of deep learning and the increasing preference for end-to-end models, differentiable simulators \cite{diff_review} have become a compelling alternative to traditional physics-based simulators such as Bullet \cite{bullet} and Mujoco \cite{todorov2012mujoco} for simulating physics. Unlike conventional simulators, these analytical differentiable simulators enable automatic differentiation, allowing seamless integration with gradient-based optimization techniques \cite{taichi, brax, JaxMD, genesis}. Several frameworks have been developed for fluid dynamics, such as PhiFlow \cite{PhiFlow}, which is framework-agnostic, JAX-Fluids \cite{JaxFluid}, built on JAX, and FluidLab \cite{fluidlab}, which leverages Taichi \cite{taichi}.  Hybrid approaches have also emerged, combining analytical equations with neural networks to improve accuracy and generalization \cite{spnet, cnn_liq, heart_neuralfluid}.

\begin{figure*}[!t]
    \centering
    \includegraphics[width=\textwidth]{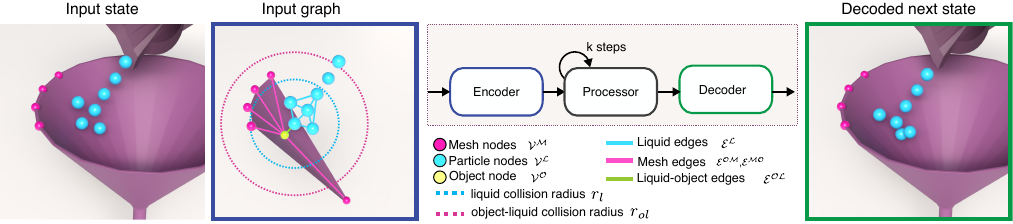}
    \caption{Overview of our network architecture and graph encoding scheme. The figure shows the graph connectivity of a liquid particle with its neighboring particles and a Cup object. Note that, for illustration purposes only, the closest point on the surface of the cup to the particle is shown as $\mathcal{V}^{O}$. In reality, the object node $\mathcal{V}^{O}$ is virtual and located at the object's center of mass.}
    \label{fig:arch}
\end{figure*}

Purely learning-based simulators, in particular, Graph Neural Networks (GNNs) \cite{sanchez2020learning, dpinet, meshgraphnet, rigid_discontinuos, fignet, sdf_learning, hopper_control_gnn, gmn, egnn, sgnn, SURFsUP} have gained significant traction due to their ability to model complex physical interactions. This success can be attributed to the inherent nature of many physical phenomena – the dynamics of rigid bodies, deformable bodies, fluids, and even their coupled interactions can often be described by local interactions between constituent entities. GNNs, with their strong relational inductive bias \cite{gnn_theory}, provide a natural and effective framework for encoding this locality, allowing the model to learn the underlying physics from data. The seminal work by \citet{sanchez2020learning} established several key architectural and representational choices that have influenced a number of subsequent works. These include using relative geometric features to define node and edge attributes, employing one-step prediction training instead of full rollout-based supervision, and utilizing message-passing GNNs \cite{messagePassingNN, interaction_network} as the core model. While this work primarily focused on particle-based fluid simulation, later studies successfully extended similar architectures to soft-body \cite{meshgraphnet} and rigid-body \cite{fignet, rigid_discontinuos, gmn, egnn, sgnn} simulations.

In physical simulation, accurately modeling collisions is essential. In GNN-based simulators, this is typically addressed through node-based neighborhood interactions. This approach has proven effective for simulating liquid-liquid \cite{sanchez2020learning} and deformable body \cite{meshgraphnet} interactions. However, it struggles when dealing with rigid-body collisions, as mesh vertices are often sparse and fail to capture fine-grained object geometry \cite{fignet, sdf_learning, SURFsUP}. To address these limitations, \citet{fignet} proposed handling collisions in the mesh-face space instead of the vertex space, leading to more robust rigid-body interactions. Another widely adopted approach is the use of SDFs for collision handling. SDFs implicitly represent surface geometry by encoding the distance of any point in space to the closest surface on a mesh. This allows for efficient collision detection by evaluating the SDF at query points. \citet{sdf_learning} used learned SDFs to simulate rigid body dynamics while \citet{SURFsUP} used SDF formulation to learn fluid dynamics in a static free-fall environment, representing all rigid surfaces in the environment as a single union of SDFs. These methods significantly reduce the computational overhead compared to node-based collision handling, offering a scalable alternative for complex physical interactions.

So far, the aforementioned GNN-based simulators have either been designed for static free-fall environments, restricting their use to design optimization tasks \cite{inverse_design, SURFsUP}, or have supported manipulation only in limited settings, typically involving simple rigid bodies like cuboids \cite{dpinet, sanchez2020learning} or spheres \cite{meshgraphnet}.  In contrast, our model extends beyond these constraints by generalizing to arbitrary scenes and complex manipulation tasks, marking a significant step toward a fully learned, general-purpose differentiable simulator capable of handling diverse physical interactions.

\section{Method}
Our learned simulator predicts the dynamics of liquid particles interacting with kinematic rigid bodies under active manipulations. Given the initial state of the liquid particles $X^0$ in terms of their positions, and external 6-DOF control inputs at every time step $U^{(0)},...,U^{(T-1)}$ for all the rigid bodies in the scene, the GNN simulator iteratively applies the learned model $X^{(t+1)} = s_\theta(X^t, U^t)$ and produces the rollout trajectory of particles as $X^{(1)},...,X^{(T)}$. The rigid bodies in our framework can either be \textit{kinematic} objects, which can be subjected to external control, or \textit{stationary} objects, that remain fixed throughout the simulation. At timestep $t$, each \textit{kinematic} object can be transformed to a new 6-DOF pose under the transformation $\mathcal{T}_t$. For \textit{stationary} objects, pose remains fixed throughout the simulation.

Our graph network follows the Encode-Process-Decode architecture introduced in \cite{sanchez2020learning}. The state of the world is first encoded into a graph where node features are derived from the current state, and edge connections are determined based on factors such as spatial proximity for potential collisions and the underlying geometry of objects. Subsequently, both node and edge features are embedded into a latent space via MLPs. The final encoded graph is passed through $P$ message-passing layers to generate a sequence of $P$ latent graphs. Finally, a decoder, which is also implemented as an MLP, extracts the updated dynamics from task-relevant nodes of the final latent graph.

\subsection{Graph Network}
We define the graph as $\mathcal{G}=(\mathcal{V}^{L}, \mathcal{V}^{O}, \mathcal{V}^{M}, \mathcal{E}^{L}, \mathcal{E}^{OL}, \mathcal{E}^{OM}, \mathcal{E}^{MO})$. In the input graph representation, $\mathcal{V}^L=\{p_i\}_{i=1 \ldots N}$ corresponds to the set of $N$ liquid particles. $\mathcal{V}^O=\{o_i\}_{i=1 \ldots Q}$ represents the set of $Q$ rigid-bodies in the scene. $\mathcal{V}^M=\{m_{ik}\}_{i=1\ldots Q,k=1 \ldots K_i}$ represents the set of surface vertices of all the objects in the scene. The surface vertices are defined such that when an object $o_i$ is transformed to a new 6-DOF pose at time $t$, the surface vertices $m_{ik}$ belonging to the object also move under the same transformation. Figure~\ref{fig:arch} gives an overview of our architecture.

Graph Connectivity in our model is handled using four sets of edges. Following \cite{fignet, sdf_learning}, we connect bi-directional edges between object node $o_i$ and its corresponding surface nodes $\{m_{ik}\}$, forming edge sets  $\mathcal{E}^{OM}$  (object-to-mesh) and  $\mathcal{E}^{MO}$ (mesh-to-object). This facilitates the propagation of object motion to the particles through the surface nodes. Interaction between liquid particles is captured using the edge set $\mathcal{E}^L$. We connect edges between liquid particles if the distance between them is less than the liquid-liquid connectivity radius $r_l$. This promotes local interactions of particles, and we choose radius $r_l$ to ensure each particle has roughly 10-20 neighbors.

Crucially, to model the interactions between fluid particles and rigid bodies, we introduce the $\mathcal{E}^{OL}$ edge set. Similar to liquid-liquid interactions, the dynamics of fluid particles are primarily affected when they are near a rigid body's surface. Notably, this interaction can occur at any arbitrary point on the object's surface, not just at mesh vertices. To achieve this, we define a function $\mathbf{C}(p'_j, o_i)$ based on the Boundary Vector Hierarchy (BVH) algorithm \cite{bvh}. This function takes the mesh geometry of object $o_i$ and a query particle $p'_j$ as input, and efficiently computes the closest point $c'_{ij}$ on the surface of the mesh to the query particle. Since this function operates on positions in the object's local frame, we first transform the liquid particle position $p_j$  into this frame using the inverse transformation $\mathcal{T}_t^{-1}(p_j)$ based on the object's current transformation $\mathcal{T}_t$. The closest point is then transformed back to the world coordinate frame:

\begin{equation} \label{eq:collision}
    c_{ij} = \mathcal{T}_t( \mathbf{C}( \mathcal{T}_t^{-1}(p_j), o_i ) )
\end{equation}

If the distance $d$ between the particle $p_j$ and the transformed closest point $c_{ij}$ is less than the liquid-object connectivity radius, $r_{ol}$, we connect an edge between the particle node and the object node. This approach is similar to \cite{sdf_learning}, where learned SDFs for collision detection between rigid bodies were used.
We use a different connectivity radius $r_{ol}$ for liquid-object interactions since our ground truth simulator maintains some separation distance between the object surface and the particles to enforce its boundary condition. 

\textbf{Node features} Following the approach of \cite{sanchez2020learning, meshgraphnet, fignet, sdf_learning, SURFsUP}, we use a history of velocity information as the main node feature. Specifically, we use the finite-difference estimates of velocity from the previous five timesteps. We calculate these velocity features differently depending on the node type. For liquid particle nodes $\mathcal{V}^L$ and mesh surface nodes $\mathcal{V}^M$, we calculate them based on their positions over the past five timesteps. For object nodes $\mathcal{V}^O$, which represent the rigid bodies, we derive velocity information from the object's 6-DOF pose at each timestep. This 6-DOF pose information is crucial for the network to learn the rotational dynamics of the objects, in addition to their translational motion. Additionally, we also add features to distinguish between different types of object nodes $\mathcal{V}^O$. Our simulation includes \textit{kinematic} objects (which are controlled) and  \textit{stationary} objects (which remain fixed). To represent this, we use a one-hot encoding to indicate its type. This one-hot encoding is also extended to the mesh nodes $\mathcal{V}^M$ since each mesh node belongs to a specific object.

\begin{figure*}[t]
    \centering
    \includegraphics[width=\textwidth]{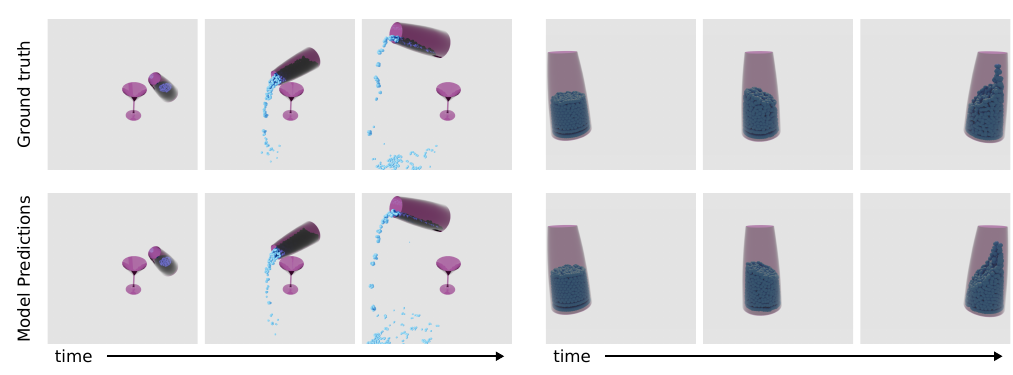}
    \caption{Example rollouts of our \textit{Full-model} on the held-out test set. }
    \label{fig:rollout}
\end{figure*}

\textbf{Edge features} For the edge sets $\mathcal{E}^L$ (liquid-liquid), $\mathcal{E}^{MO}$ (mesh-to-object), and $\mathcal{E}^{OM}$ (object-to-mesh), we adopt the relative encoding technique \cite{sanchez2020learning}. The edge features consist of the relative displacement vector between two connected nodes and the scalar distance between them. The calculation of these relative features differs slightly depending on the edge type. For $\mathcal{E}^L$, the relative position and distance are computed between the positions of the two connected liquid particles. For $\mathcal{E}^{MO}$ and $\mathcal{E}^{OM}$ edges, we calculate them based on the mesh node position and the center of mass position of the corresponding object, derived from the object node's 6-DOF pose. For the liquid-object edge set $\mathcal{E}^{OL}$, which is critical for modeling particle collisions with surfaces, we use a specialized feature set. For an edge connected between particle $p_j$ and object $o_i$, the feature vector is defined as $e_{ij}^{ol}= [c_{ij} - p_{j}, c_{ij} - p_{o_i}, \lVert c_{ij} - p_{j} \rVert, \lVert c_{ij} - p_{o_i} \rVert]$. Here, $c_{ij}$ represents the closest point on the surface of object $o_i$ to the particle $p_j$, and $p_{o_i}$ represents the center of mass of object $o_i$.  These features together give information about the spatial relationship between the connected particle and object, using the closest point on the object's surface $c_{ij}$, as a reference. The resulting input graph is encoded into separate MLPs for each node type and edge type before passing to the processor block.

\textbf{Processor} The processor stage consists of $P$ identical message-passing steps, each with its own learned parameters. These blocks iteratively update the embedded node and edge features using the following update equations:
\begin{align}
    e_{ij}^{\prime L} &= f^{LL}(e_{ij}^L, \mathbf{v}_i^L, \mathbf{v}_j^L) \label{eq:ll} \\
    e_{ij}^{\prime OL} &= f^{OL}(e_{ij}^{OL}, \mathbf{v}_i^O, \mathbf{v}_j^L) \label{eq:ol} \\
    e_{ij}^{\prime MO} &= f^{MO}(e_{ij}^{MO}, \mathbf{v}_i^M, \mathbf{v}_j^O) \label{eq:mo} \\
    e_{ij}^{\prime OM} &= f^{OM}(e_{ij}^{OM}, \mathbf{v}_i^O, \mathbf{v}_j^M) \label{eq:om} \\
    \mathbf{v}_i^{\prime M} &= f^M(\mathbf{v}_i^M, \textstyle\sum_{j} e_{ij}^{\prime OM}) \label{eq:m} \\
    \mathbf{v}_i^{\prime O} &= f^O(\mathbf{v}_i^O, \textstyle\sum_{j} e_{ij}^{\prime MO}) \label{eq:o} \\
    \mathbf{v}_i^{\prime L} &= f^L(\mathbf{v}_i^L, \textstyle\sum_{j} e_{ij}^{\prime L}, \textstyle\sum_{j} e_{ij}^{\prime OL}) \label{eq:l}
\end{align}
where $f^{LL}$, $f^{OL}$, $f^{MO}$, $f^{OM}$, $f^{M}$, $f^{O}$, $f^{L}$ are each implemented as MLPs.

\textbf{Decoder} The processor updates the latent representations of all the node types. However, since our goal is to predict particle dynamics under the influence of rigid body manipulations, the decoder, implemented as an MLP, only decodes latent features of liquid nodes $\mathcal{V}^L$. We train the model using the $L2$ loss on one-step acceleration predictions of the liquid particles. Additional details on the network architecture and training procedure are provided in Appendix \ref{Arch}.

\begin{figure*}[t]
    \centering
    \includegraphics[width=\textwidth]{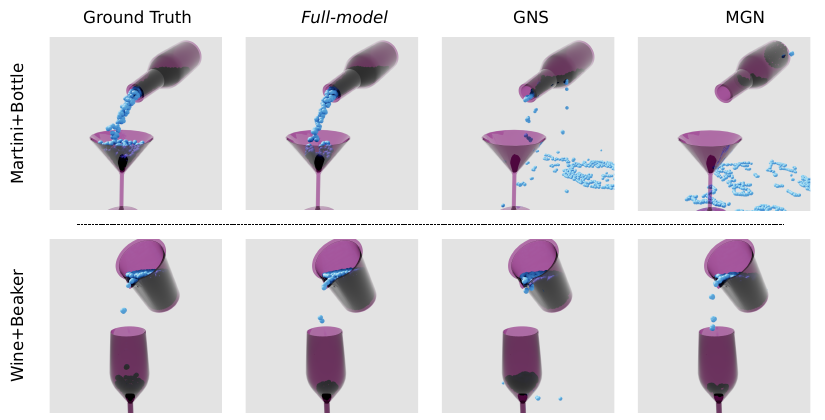}
    \caption{Comparison with the baseline models. GNS and MGN struggle to handle particle dynamics with novel geometry, while our method generalises without any additional training.}
    \label{fig:gns_compare}
\end{figure*}

\section{Experiments}

\subsection{Ground Truth Simulator} We use NVIDIA Isaac Sim \cite{isaac_sim} as our ground truth simulator. Isaac Sim is a robotics simulation platform that also supports particle simulation based on the Position Based Dynamics (PBD) \cite{pbd} method. To investigate the ability of our system to generalize from a small set of training data, our primary dataset consisted of a scene with a collection of liquid particles and three rigid bodies: a jug (\textit{kinematic}), a cup (\textit{stationary}), and a floor (\textit{stationary}). We adjusted the properties of particles to approximate those of water. Each simulation episode was initialized as follows: the cup was placed at a random location, the floor was positioned underneath it, and the jug was placed above the floor, containing the liquid particles. To enable the network to learn liquid dynamics  under a wide range of jug motions, as well as the effects of liquid collisions with stationary objects (cup and floor), we create three types of simulation scenarios:
\begin{itemize}
    \item \textbf{Translation-Motion-Sim}:  The jug undergoes purely translational motion, with a random velocity along the $x$, $y$, or $z$ axis at each timestep. This scenario teaches the network how linear motion of a container affects the enclosed liquid.
    \item \textbf{Rotation-Motion-Sim}:  The jug remains positionally fixed but rotates with a random angular velocity around a random axis. In half the cases, the rotation is configured to pour liquid into the cup, and in the other half, the liquid spills on the floor. This helps the model learn the effects of rotational motion on the liquid, as well as collisions with the stationary objects.
    \item \textbf{Full-Body-Motion-Sim}: The jug undergoes combined translational and rotational motion. It moves upwards along a specified direction in the $xy$-plane while simultaneously rotating with a random angular velocity. This motion is designed to simulate a pouring action and allows the network to learn complex fluid dynamics resulting from 6-DOF motion.
\end{itemize}

To train the network to handle abrupt changes in liquid motion, we introduce random noise to the jug's motion in 30\% of the trials. We use a \textit{Martini} cup and \textit{Vase} jug to generate the dataset for training our model. Our training and test sets contain 1200 and 120 simulations(each with 415 timesteps), respectively.

\textbf{Generalization Test Sets} To evaluate our model's generalization capabilities, we created additional \textit{Rotation-Motion-Sim} simulations using two novel jug shapes, \textit{WineBottle} and \textit{Beaker}, and a \textit{Wine} cup. Our training dataset used the \textit{Vase} jug, which has a relatively simple geometry. In contrast, the \textit{WineBottle} has a wide base that tapers sharply to a narrow opening. The \textit{Beaker} has a spout at the rim, channeling the liquid into a narrower stream. We also used a \textit{Wine} cup, which features a curved geometry, unlike the conical shape of the \textit{Martini} cup used during training.

\subsection{Evaluation}
\textbf{Baseline models} We compare our model against recent GNN-based particle simulators that support kinematic rigid body manipulation, specifically Graph Network Simulator (GNS) \cite{sanchez2020learning} and MeshGraphNet \cite{meshgraphnet}. Both models handle liquid-object interaction using node-based collision. GNS uses a single edge set to model both liquid-liquid and liquid-object interactions. In contrast, MeshGraphNet separates liquid-liquid and liquid-object interactions into two distinct edge sets. We evaluate two variants of MeshGraphNet based on how the liquid-object edge set is defined.  In the first variant (MGN), it includes only liquid-object edges. In the second variant (MGN*), mesh connectivity edges are included alongside the liquid-object edges. All 3 models use a single node set to represent both particle and mesh nodes. As a result, the 6-DOF control inputs for manipulating a rigid body are incorporated into the node features of all the nodes in the graph. This implicit representation of control inputs limits these models to manipulating only one rigid object per simulation. We use the same neighborhood radii as in our model: $r_l$ for liquid-liquid collision and $r_{ol}$ for liquid-object collision. See Appendix \ref{baseline} for complete details of the baseline models.

\textbf{Ablation} We hypothesized that for our task, the mesh vertex nodes $\mathcal{V}^M$  might be less critical for learning liquid-object interaction, and that the direct interaction between $\mathcal{V}^L$ and $\mathcal{V}^O$ via our accurate collision handling method might suffice. While the mesh nodes $\mathcal{V}^M$ do contribute to updating latent states of object nodes $\mathcal{V}^O$ within the processor block, their direct influence on the liquid nodes $\mathcal{V}^L$ might be limited. To test this, we created an ablated version of our model by removing the mesh vertex nodes $\mathcal{V}^M$ and the corresponding edge sets $\mathcal{E}^{OM}$ and $\mathcal{E}^{MO}$. This results in a simplified input graph: $\mathcal{G}=(\mathcal{V}^{L}, \mathcal{V}^{O}, \mathcal{E}^{L}, \mathcal{E}^{OL})$. The processor update equations are reduced to Equations~\eqref{eq:ll}, \eqref{eq:ol}, and \eqref{eq:l}, with Equation~\eqref{eq:o} simplifying to $\mathbf{v}_i^{\prime O} = \mathbf{v}_i^O$ (meaning the object node features are not updated within the processor block). This ablated model has a lower memory footprint during both training and inference. In the rest of the paper, we refer to our complete network as \textit{Full-model} and to this reduced version as \textit{Ablated-model}. The baseline models and the \textit{Ablated-model} were trained using the same dataset as our proposed model.

\textbf{Metric} For quantitative comparison, we use the mean Chamfer distance over the entire rollout trajectory between the ground truth and the model prediction. Chamfer distance is more suitable for evaluating point cloud data than root mean squared error (RMSE), as it emphasizes overall distributional similarity rather than exact one-to-one correspondences between particles.

\begin{figure*}[t]
    \centering
    \includegraphics[width=\linewidth]{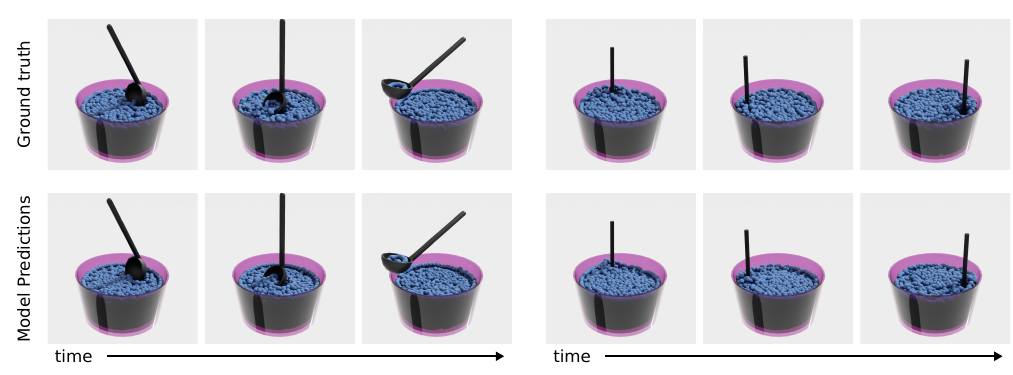}
    \caption{Rollouts of our \textit{Full-model} showing successful  \textit{Scooping} (left) and \textit{Stirring} (right) manipulation tasks} 
    \label{fig:manipulation}
\end{figure*}

\subsection{Results}
Our graph network model reliably predicts particle dynamics under external manipulations of rigid bodies. We first evaluate the \textit{Full-model}'s long-term rollout performance on the test set by comparing the predicted rollouts with the ground truth simulator, qualitatively assessing the model's ability to maintain realistic motion over extended periods. During rollouts, the ground truth 6-DOF control inputs for the jug are provided to the network at each timestep. Figure~\ref{fig:rollout} presents qualitative results for our \textit{Full-model}, showing rollout snapshots from two challenging scenarios. In the \textit{sliding} task (Figure~\ref{fig:rollout} (right) from the \textit{Rotation-Motion-Sim} test set), when the externally controlled jug comes to an abrupt stop, our model accurately handles both the boundary collisions and the conservation of momentum of the liquid particles. In the \textit{full-motion} task (Figure~\ref{fig:rollout} (left) from the \textit{Full-Body-Motion-Sim}), the jug is manipulated across all six degrees of freedom. Our network successfully tracks the intricate dynamics throughout the rollout. Notably, on these test sets, both the \textit{Ablated-model} and the GNS baseline exhibit qualitatively similar performance to the \textit{Full-model}, whereas the MGN and MGN* models struggle to maintain consistency over the entire rollout.

\begin{figure}[h!]
    \centering
    \includegraphics[width=0.8\linewidth]{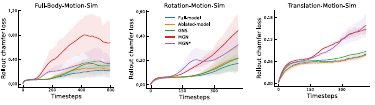}
    \caption{Mean Chamfer loss curves over rollout on the three test sets for all models. Our model achieves the lowest error; shaded regions show variability as median absolute deviation.}
    \label{fig:err_traj}
\end{figure}

\textbf{Generalization}  A key advantage of our model is its ability to generalize to novel object shapes. Figure~\ref{fig:gns_compare} compares rollouts from our \textit{Full-model}, GNS, and MGN on the Wine-Beaker and Martini-Bottle test sets. Despite never encountering these shapes during training, our model accurately simulates the liquid dynamics, closely matching the ground truth behavior. Importantly, our \textit{Ablated-model} also performs comparably well, suggesting that both models effectively learn to represent local surface geometry through our liquid-object collision handling. In contrast, the GNS and MGN baselines exhibit substantial limitations. With the \textit{Beaker}, GNS shows pronounced tunneling effects, with particles incorrectly passing through the spout walls. While MGN is able to channel particles through the spout, it still suffers from tunneling toward the end of the rollout. With the \textit{WineBottle}, both GNS and MGN struggle to simulate the liquid flow as the particles approach the narrow neck, resulting in significant particle tunneling. This difficulty stems from their reliance on mesh vertices alone to handle liquid-object collisions, which fails to capture finer geometric details of these novel shapes. Importantly, this is despite increasing the mesh vertex density of the objects by 2-5 times to facilitate better shape information. MGN* performs worse than both GNS and MGN across all evaluated scenarios. Table~\ref{tab:results} summarizes the quantitative results for each model, while Figure~\ref{fig:err_traj} shows the step-wise error over the trajectory. Our \textit{Full-model} consistently achieves the best scores across most test scenarios. Additional snapshots can be found in Appendix \ref{snaps}.

\begin{table}
\caption{Quantitative comparison of \textit{Full-model}, \textit{Ablated-model}, and the baseline models using mean Chamfer Distance (\textbf{lower is better}) with respect to ground truth simulations. $N$ represents the number of particles. All simulations had a rollout length of 415 timesteps.}
\centering
\vspace{1em} 
\begin{tabular}{@{}lcccccc@{}}
\toprule
Simulation & \multirow{2}{*}{$N$} & \textit{Full-} & \textit{Ablated-} & \multirow{2}{*}{GNS} & \multirow{2}{*}{MGN} & \multirow{2}{*}{MGN*} \\
 domain & & \textit{model} & \textit{model} & & & \\
 
\midrule

Translation- & \multirow{2}{*}{1K} & \multirow{2}{*}{0.057$\pm$0.01} & \multirow{2}{*}{\textbf{0.055$\pm$0.01}} & \multirow{2}{*}{0.063$\pm$0.02} & \multirow{2}{*}{0.092$\pm$0.06} & \multirow{2}{*}{0.089$\pm$0.04} \\
Motion-Sim & &  &  &   &  &\\

Rotation- & \multirow{2}{*}{1K} & \multirow{2}{*}{\textbf{0.086$\pm$0.03}} & \multirow{2}{*}{0.099$\pm$0.05} & \multirow{2}{*}{0.099$\pm$0.04} & \multirow{2}{*}{0.178$\pm$0.18} & \multirow{2}{*}{0.163$\pm$0.07} \\
Motion-Sim & &  &  &   &  & \\

Full-Body- & \multirow{2}{*}{1K} & \multirow{2}{*}{\textbf{0.138$\pm$0.07}} & \multirow{2}{*}{0.165$\pm$0.09} & \multirow{2}{*}{0.184$\pm$0.09} & \multirow{2}{*}{0.493$\pm$0.37} & \multirow{2}{*}{0.236$\pm$0.14} \\
Motion-Sim & &  &  &  &  &  \\

Wine-Beaker                 & 1K   & \textbf{0.091$\pm$0.02} & 0.111$\pm$0.035 & 0.156$\pm$0.01 & 0.141$\pm$0.04 & 0.347$\pm$0.06\\
Martini-Bottle              & 1K   & 0.077$\pm$0.01 & \textbf{0.069$\pm$0.01} & 2.02$\pm$0.15 & 1.317$\pm$0.15 & 2.261$\pm$0.17\\

\bottomrule
\end{tabular}

\label{tab:results}
\end{table}

We further validated the generalization capabilities of our \textit{Full-model} by simulating liquid pouring from the classic Utah teapot, which features intricate geometry. As shown in Figure~\ref{fig:2jugs_tea}, the \textit{Full-model} (chamfer loss of 0.078) successfully simulates realistic liquid dynamics even for this challenging shape. The \textit{Ablated-model} (chamfer loss of 0.301), however, exhibited some leakage with the Utah teapot, suggesting a potential limitation in handling highly complex geometries.  
The difference in performance may be due to the unavailability of object node $\mathcal{V}^O$ updates during the message-passing process in the \textit{Ablated-model}. 

In practice, since both the \textit{Full-model} and \textit{Ablated-model} offer similar performance in most scenarios, the two can be seen as complementary. The \textit{Ablated-model} is suitable for environments with rigid objects of relatively simple shapes, as it reduces computational overhead and memory footprint, while the \textit{Full-model} is preferable for scenes containing highly intricate geometries.

\subsection{Novel Manipulation Scenarios} 
Having demonstrated in the previous section that our network generalises to novel objects, we now investigate our model's ability to handle unseen manipulation tasks. We focus on \textit{Stirring} and \textit{Scooping} tasks, two common liquid manipulation tasks found in kitchen settings that are significantly different from the pouring motions used during training. We created a new simulation environment consisting of a large pot filled with liquid particles (2.1K particles). In the \textit{Stirring} task, a stick is controlled to stir the liquid. In the \textit{Scooping} scenario, a ladle is manipulated to scoop liquid from the pot. We compare the performance of our \textit{Full-model} against ground truth simulations generated by Isaac Sim. Figure~\ref{fig:manipulation} illustrates the trajectories of these two manipulation tasks performed by our \textit{Full-model} along with the ground truth. In both the manipulation tasks, our model successfully simulates the complex liquid dynamics. For the \textit{Stirring task} (chamfer loss of 0.0514), the network accurately models the pushing forces exerted by the stick on the liquid particles as it rotates. For the \textit{Scooping task} (chamfer loss of 0.0593), the model captures both the scooping action and the subsequent containment and transport of the liquid within the ladle. Additionally, we also showcase our model's ability to simultaneously control multiple objects. We created a two-jug pouring scenario, where each jug is independently controlled to pour liquid. Figure~\ref{fig:2jugs_tea} shows the snapshots of the result from our \textit{Full-model} (chamfer loss of 0.257), where it realistically simulates the interaction of liquid streams from two sources.  Taken together, these results show that our learned simulator is indeed a versatile general-purpose simulator for modeling liquid dynamics. It accurately captures complex interactions between liquids and rigid bodies of arbitrary shapes undergoing diverse manipulations.

\subsection{Control on learned dynamics}

A key advantage of a fully differentiable environment is the ability to solve tasks using gradient-based optimization techniques. To showcase the utility of our model, we designed a pouring task where the objective is to pour a specified percentage of the cup's total volume from the jug into the cup. In the initial state, the jug is placed in an upright position next to the cup, such that rotating the jug around the $x$-axis will cause liquid to pour into the cup. We use an MPC controller with the control variable being the jug's rotational acceleration angle around the $x$-axis. The control inputs are initialized randomly. Following a task formulation similar to \cite{spnet}, we implement a two-stage cost function optimized using Adam \cite{ADAM}. In the first stage, the cost is the $L2$ distance between all liquid particles and a target point on the rim of the jug closest to the cup. This encourages the jug to rotate towards the cup to let the liquid flow into the cup. Once the target fill level is reached, the second stage activates. This stage uses a regularization term on the jug's rotation angle, encouraging it to return to its initial upright position.

\begin{wrapfigure}{r}{0.44\textwidth} 
  \vspace{-4pt}
  \centering
  \includegraphics[width=\linewidth]{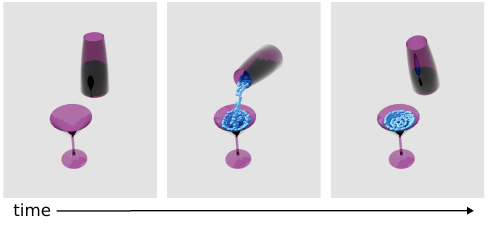}
  \caption{Optimal trajectory snapshots from our MPC controller for a 30\% fill target.}
  \label{fig:mpc}
  \vspace{-6pt}
\end{wrapfigure}

Figure~\ref{fig:mpc} shows snapshots of the optimized trajectory for a target fill level of 30\%. Despite the random initialization of control inputs, the optimizer successfully finds a control sequence that achieves the desired fill level. We tested the controller with four different target fill levels, corresponding to 10\%, 25\%, 50\%, and 70\% of the cup’s volume, and observed achieved fill levels of 10\%, 44\%, 72\%, and 94\%, respectively. For higher fill targets, there is a larger deviation from the desired level, likely due to the dynamics of the two-stage control: at higher fill targets, the jug rotates more in the first stage. When the second stage activates and the jug begins to return to its upright position, residual liquid continues to pour, leading to overfilling.

\subsection{Computational Complexity and Inference Times}\label{complexity}
For liquid-object collisions, our BVH-based handling reduces complexity compared to node-based baselines. BVH is relatively inexpensive, with a complexity of $\mathcal{O}(\log (K_i))$ per query, where $K_i$ is the number of vertices in the mesh. In our method, each particle is connected to at most one edge per object, yielding an $\mathcal{O}(NQ)$ complexity for $N$ liquid particles and $Q$ objects. In node-based collision, such as in GNS, a particle may connect to many vertices, resulting in $\mathcal{O}(NK)$ complexity, where $K$ is the total number of object mesh vertices. Since the number of objects is significantly lower than the total number of node vertices($Q << K$), a substantial speedup is possible.

\begin{table}[ht]
    \centering
    \caption{Comparison of Inference Times across all models.}
    \label{tab:step_times}
    \begin{tabular}{lc}
        \toprule
        \textbf{Method} & \textbf{Time per step (ms)} \\
        \midrule
        Full-model & 25.05 \\
        Ablated-model & 21.34 \\
        GNS & 75.51 \\
        MGN & 50.32 \\
        MGN* & 49.58 \\
        Isaac-Sim (equiv.) & 10.20 \\
        \bottomrule
    \end{tabular}
\end{table}

\begin{figure}[h]
    \centering
    \includegraphics[width=0.8\linewidth]{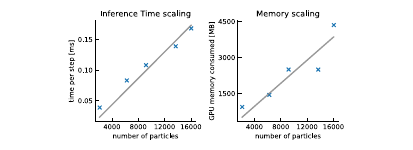}
    \caption{ Inference time and memory consumption plotted against the number of particles for our model.}
    \label{fig:memory}
\end{figure}

We calculated the rollout inference times per step ($dt=1/60$) for all our models and baselines on the test set (evaluated on a workstation with Nvidia 4090 GPU and 32-core CPU) and report them in Table~\ref{tab:step_times}. Indeed, the performance is twice as fast as the fastest baseline methods. However, it is still slower than Isaac-sim. This gap arises from CPU-bound neighborhood computations, a known limitation in these GNN-based methods, as noted in the GNS (supplementary).

For the overall system, our method scales linearly with the number of particles $\mathcal{O}(N)$. We tested this on the stirring environment by varying the number of particles between 2k and 16k particles and report the memory consumption and inference time per step in Figure~\ref{fig:memory}. The results confirm that our model scales linearly with the number of particles.
\section{Discussion}

We presented a unified GNN-based framework for learning particle-based fluid dynamics under rigid body manipulations. The key insights driving our approach are (1) separating liquid-liquid and liquid-object interactions, (2) using accurate surface-based collision handling for liquid-object interactions, (3) representing rigid bodies and fluid particles with separate node sets, making the architecture modular and extensible. These design choices enable our model to simulate fluid behavior in complex environments, setting it apart from prior work.

While promising, our current framework is limited in its ability to simulate fluids with diverse physical properties, such as viscosity. Future work will focus on incorporating richer representations of material properties within the network architecture. Additionally, our framework is currently designed exclusively for learning fluid dynamics and, as such, cannot handle rigid body dynamics when interacting with particles. For instance, it cannot simulate a scenario where flowing water displaces or carries a rigid object. However, given our modular graph definition, it is possible to extend our framework to handle this two-way coupling. By introducing trainable non-kinematic rigid-body nodes ($\mathcal{V}^R$) alongside liquid nodes ($\mathcal{V}^L$), and defining explicit liquid–object ($\mathcal{E}^{LO}$) and object–object ($\mathcal{E}^{OO}$) edges for bidirectional interactions, the framework could learn coupled fluid–rigid dynamics. We consider this a natural extension for future work.

\clearpage

\section*{Acknowledgments}
This research was supported by the European Research Council (ERC; Consolidator Award ‘ACTOR’-project number ERC-CoG-101045783), by the Hessian Ministry of Higher Education, Research, Science and the Arts with its LOEWE research priority program ‘WhiteBox’, and by ‘The Adaptive Mind’, funded by the Excellence Program of the Hessian Ministry of Higher Education, Science, Research and Art.





\bibliographystyle{unsrtnat}
\bibliography{reference}

\clearpage

\appendix

\section{Appendix}
\subsection{Architecture and training}\label{Arch}

All the MLPs in our architecture contain 2 hidden layers of 128 units each with LayerNorm \cite{layernorm} and an output layer of 128 units. Only in the case of the decoder, the output layer has 3 units to decode the 3D acceleration of particles. We chose $P=10$ as the number of processor blocks.

It has been shown that perturbing the inputs with noise makes the GNN models more robust to noisy inputs \cite{sanchez2020learning}. This is particularly helpful when generating rollouts where the network is fed with its own noisy, previous predictions as input. We choose a noise value $\mathcal{N}(0, \sigma=0.00067)$. Additionally, all the inputs and outputs were normalised.

We use Adam \cite{ADAM} to optimize the $L2$ loss on one-step acceleration predictions of the liquid particles with an exponentially decaying learning rate from 1e-4 to 1e-6. The predicted particle accelerations are then integrated to produce the next-step positions using the finite-difference method.

The models were trained on a single Nvidia 4090 GPU up to a maximum of $2 M$ gradient steps with a batch size of 20. The training period lasted approximately 10 days. The source code and visualizations are available here.\footnote{https://github.com/RothkopfLab/fluid-manip-gnn}

\subsection{Baseline model details}\label{baseline}
GNS baseline has a graph of the form $\mathcal{G}=(\mathcal{V}, \mathcal{E})$, while MeshGraphNet (MGN and MGN*) models use a multigraph of the form $\mathcal{G}=(\mathcal{V}, \mathcal{E}^{L}, \mathcal{E}^{M})$. For all three baselines, the node set $\mathcal{V}$ contains both liquid particles and rigid bodies. To represent a rigid body, its mesh vertices are used as nodes in a graph. To distinguish between particle nodes and mesh nodes, one-hot encoding is incorporated into the node features. Unlike our model, GNS does not explicitly represent the floor as a separate object; instead, the distance from each particle to the floor is included as part of each particle's node features. We use the same floor representation for MeshGraphNets as well. 

For the edge sets in MeshGraphNet baselines, $\mathcal{E}^{L}$ captures liquid–liquid interactions (within neighborhood radius $r_l$) and corresponds to $\mathcal{E}^{L}$ from our model. The $\mathcal{E}^{M}$ edge set represents liquid–object interactions, where an edge is connected between a liquid particle and a mesh vertex node if they are within the collision radius $r_{ol}$. The MGN* model additionally includes edges between surface nodes of the rigid bodies in the $\mathcal{E}^{M}$ set. In the GNS baseline, $\mathcal{E}$ edge set contains both liquid-liquid and liquid-object edges.

The architecture of the individual MLPs in the baseline models, as well as the training procedure, normalisation, and noise injection, remains the same as in our model training.

Since collision with particles is handled at the mesh vertex level in the baseline models, a low vertex count negatively impacts performance. To ensure a fair comparison, we oversampled all objects used in the baseline evaluations. Table~\ref{tab:nodes_edges} summarizes the number of nodes and edges used in both our models and the baseline models.

\begin{table}[h]
  \caption{Number of vertices and edges of each object used in our models and baselines. }
  \label{tab:nodes_edges}
  \centering
  \begin{tabular}{lcccc}
    \toprule
    \multirow{2}{*}{Object} & \multicolumn{2}{c}{Our models} & \multicolumn{2}{c}{Baseline models} \\
    \cmidrule(r){2-3} \cmidrule(r){4-5}
                           & \# Nodes & \# Edges & \# Nodes & \# Edges \\
    \midrule
    Vase jug               & 776   & 3264   & 1684   & 8640   \\
    Beaker jug              & 540   & 3264   & 1994   & 10728   \\
    WineBottle jug          & 713   & 2880   & 2261   & 10944   \\
    Martini cup                & 491   & 1800   & 909   & 4080   \\
    Wine cup             & 903   & 4032   & 2780   & 14208   \\
    \bottomrule
  \end{tabular}
\end{table}

\begin{figure}[h]
    \centering
    \includegraphics[width=1\linewidth]{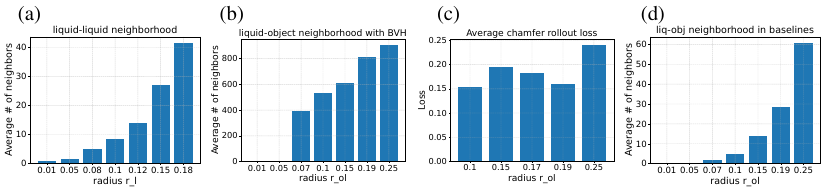}
    \caption{Effects of varying $r_l$ and $r_{ol}$ on edge connectivity and model performance.}
    \label{fig:radii}
\end{figure}

\textbf{Connectivity radius} $r_{l}$ was chosen such that each liquid particle interacts with approximately 10-20 particles. To determine this value, we computed the average number of neighbors for each particle in a subset of our main dataset (Martini cup + Vase jug) across various values of $r_l$.  Figure~\ref{fig:radii}(a) illustrates these results, based on which we selected $r_l = 0.12$.

$r_{ol}$ defines the neighborhood radius for liquid-object interactions. In Isaac Sim, a small separation is maintained between liquid particles and object surfaces to enforce boundary conditions during collision handling. As a result, particles do not interact with objects exactly at their surface. To determine the minimum value of $r_{ol}$ at which the object surface falls within the particle interaction range, we used the same dataset as for $r_l$ and ran our BVH-based collision algorithm.  Figure~\ref{fig:radii}(b) shows the average number of particles in contact with object surfaces for different values of $r_{ol}$. For values of $r_{ol}$ below 0.05, the object surface lies outside the liquid-object interaction radius. 

To identify the appropriate value for $r_{ol}$, we trained our \textit{Full-model} across different $r_{ol}$ values and observed that performance remained largely consistent between 0.1 and 0.2 (see Figure~\ref{fig:radii}(c)). For the baseline models, which use node-based collision detection, we ran a similar test as for $r_l$ (Figure~\ref{fig:radii}(d)). We ultimately selected $r_{ol} = 0.19$ for all experiments, as it ensures that each liquid particle is connected to approximately 20-30 object mesh nodes in the baseline models, allowing a fairer comparison with our models.

\subsection{Failure cases} 
We tested our model on two scenarios where it fails or performs suboptimally. 

Tunneling occurs when object velocities are significantly higher than those seen during training. GNN-based models operate on a fixed time step (dt=1), and hence, if the object pose changes significantly between steps, the object can jump across particles between steps, resulting in missed collisions (“tunneling”). We tested this on the Martini-Vase environment (identical to our training environment) by making the Vase jug rotate rapidly. We observe the tunneling effect at the beginning of the simulation (see video in our project repository). In Classical simulators, this tunneling effect is typically mitigated by using sub-stepping and continuous collision detection (CCD) methods. In learned GNN-based simulators, it is not clear how such remedies can be integrated.

Another case where our model fails is when the liquid motion is caused by shear stress and centrifugal effects. For example, when a cylindrical container with liquid inside spins about its axis, the liquid also rotates along with the container. This behavior cannot be captured by our model, as our liquid–object connectivity is distance-based. When the container rotates around its axis, particle–surface distances remain constant, so the model does not perceive tangential motion. To test this, we created a cement-mixer environment in which liquid particles are placed inside a rotating drum. Our model fails to reproduce the expected rotation of the liquid (see video in our project repository).

\clearpage

\subsection{Additional Figures}\label{snaps}

\begin{figure}[h]
    \centering
    \includegraphics[width=.85\linewidth]{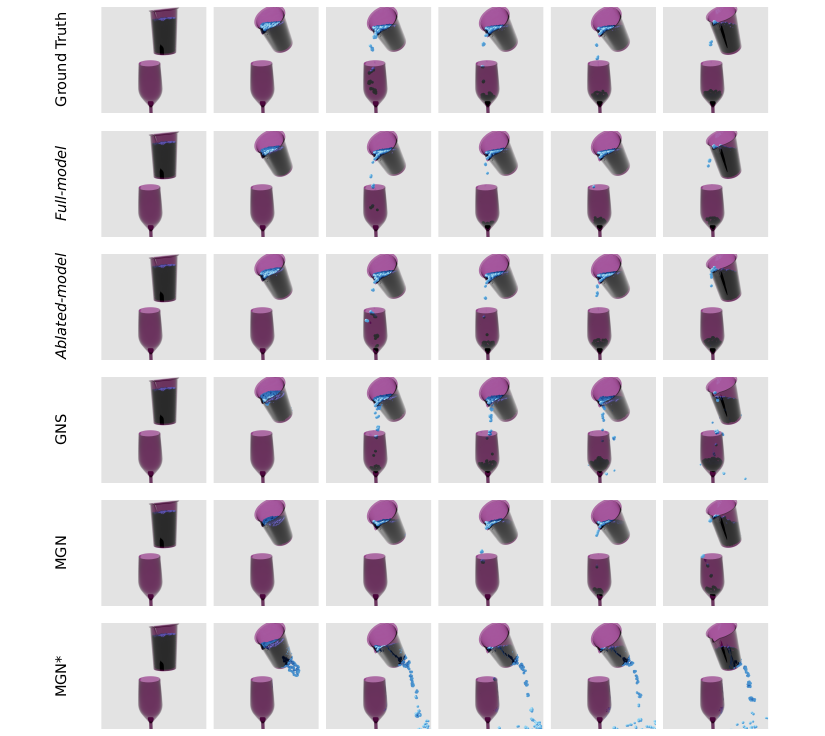}
    \caption{Visualisation of rollouts of all models on a Wine+Beaker example}
    \label{fig:wineBeak_supp}
\end{figure}

\begin{figure}[h]
    \centering
    \includegraphics[width=.8\linewidth]{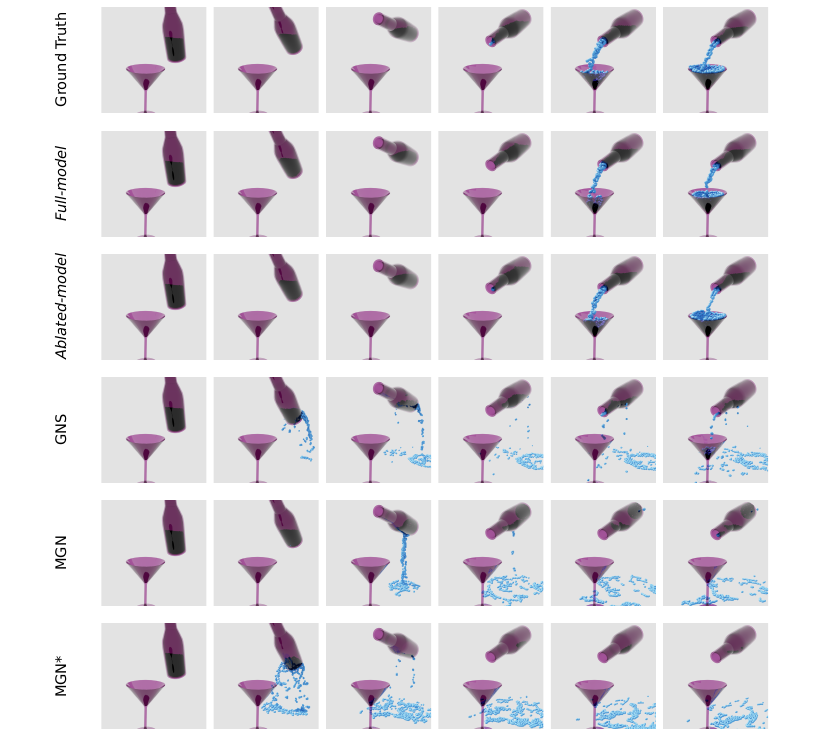}
    \caption{Visualisation of rollouts of all models on a Martini+Bottle example}
    \label{fig:martBot_supp}
\end{figure}

\begin{figure}[h]
    \centering
    \includegraphics[width=.8\linewidth]{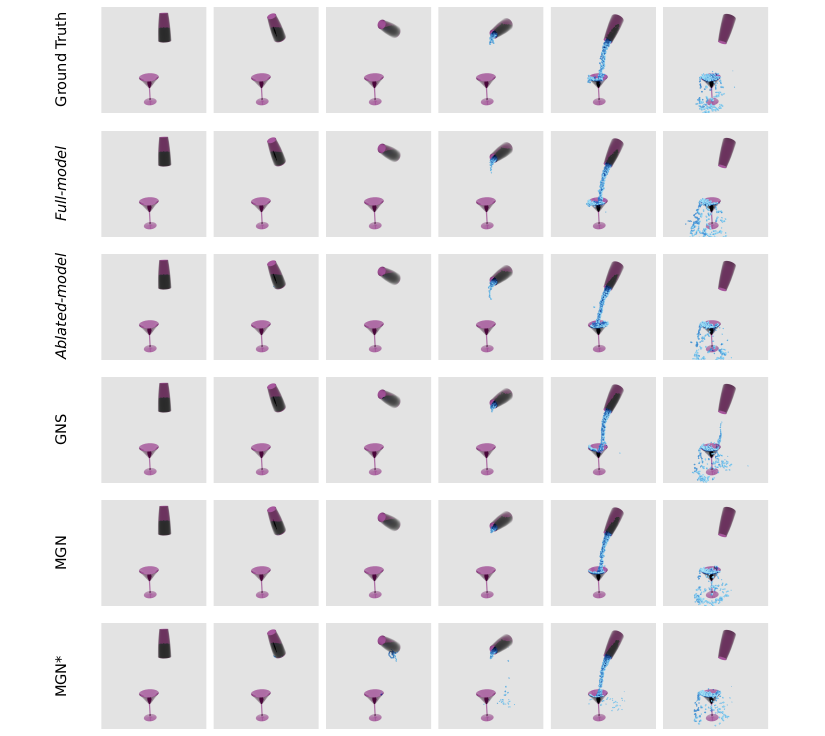}
    \caption{Visualisation of rollouts of all models on a Martini+Vase example}
    \label{fig:martVase_supp}
\end{figure}


\end{document}